\documentclass[runningheads]{llncs}
\usepackage[T1]{fontenc}
\usepackage{graphicx,verbatim,amsmath}
\usepackage{color,soul}
\usepackage{physics}
\usepackage{booktabs,tabularx,graphicx,array}
\usepackage{algorithm}
\usepackage{booktabs,tabularx,graphicx,array,subcaption,enumitem}
\usepackage{amssymb}
\usepackage{hyperref}

\newcommand{\be}{\begin{equation}}
\newcommand{\ee}{\end{equation}}

\begin{document}
\title{ImPartial: Multi-channel Whole-Cell Segmentation using Partial Annotations}
\titlerunning{ImPartial}
\author{Gunjan Shrivastava \and
Saad Nadeem}
%index{Shrivastava, Gunjan}
%index{Nadeem, Saad}
%
\authorrunning{Shrivastava et al.}
\institute{Memorial Sloan Kettering Cancer Center, New York NY 10065, USA
\email{nadeems@mskcc.org}}
  
\maketitle              % typeset the header of the contribution
\begin{abstract}{Accurate cell segmentation in pathology images typically requires dense pixel-wise annotations, which are costly and time-consuming to obtain. This challenge is especially important for emerging biological imaging modalities and multiplexed datasets with variable channel configurations, where expert-labeled data are scarce. In this work, we introduce ImPartial, a deep learning framework designed to achieve state-of-the-art segmentation performance in low-annotation regimes using sparse scribbles and limited supervision. ImPartial augments the segmentation objective via self-supervised multi-channel quantized imputation. This approach leverages the observation that perfect pixel-wise reconstruction or denoising of the image is not needed for accurate segmentation, and thus, introduces a self-supervised classification objective that better aligns with the overall segmentation goal. We demonstrate that ImPartial achieves performance at par with fully supervised models while requiring substantially fewer annotations. Extensive experiments on benchmark multiplexed cellular imaging and single-plex clinical brightfield immunohistochemistry datasets show consistent improvements over strong baselines with only partial annotations. All benchmark datasets and code are available via our Github:  \href{https://github.com/nadeemlab/ImPartial}{https://github.com/nadeemlab/ImPartial}.}

% Authors must provide keywords and are not allowed to remove this Keyword section.

\end{abstract}

\section{Introduction}

Accurate cell segmentation remains a fundamental yet challenging problem in biomedical image analysis, particularly for emerging microscopy and highly multiplexed spatial imaging modalities where annotated data are scarce, channel dimensionality varies widely (e.g., 1–100+ channels), and image quality is variable. State-of-the-art methods such as Cellpose \cite{pachitariu2022cellpose}, Mesmer \cite{greenwald2022whole}, Cellotype \cite{pang2024cellotype}, and CellSAM \cite{Marks2025-xd} have improved robustness and cross-dataset generalization through large curated training corpora, architectural inductive biases (e.g., flow-based representations), or adaptation of foundation models such as Segment Anything Model \cite{kirillov2023segment}. However, these approaches typically rely on dense, high-quality pixel-wise annotations and assume relatively consistent channel configurations (e.g. two nuclei and membrane channels). In practice, many biological datasets contain only partial annotations (e.g., incomplete masks or sparse scribbles), variable number of channels, substantial imaging noise, staining variability, low signal-to-noise ratios, and pronounced domain shifts across acquisition platforms, limiting their applicability in low-data regimes.

To alleviate annotation burden and improve robustness to noise, self-super\-vised and joint denoising–segmentation approaches have been proposed. Blind-spot denoising frameworks such as Noise2Void \cite{krull2019noise2void} and joint models such as DenoiSeg exploit complementary restoration and segmentation objectives to enhance performance with limited supervision. In particular, DenoiSeg \cite{buchholz2020denoiseg} extends blind-spot training to simultaneously learn image restoration and segmentation from partially labeled data. Nevertheless, these methods generally assume access to reasonably reliable masks for a non-trivial subset of training samples and are primarily designed for pixel-independent noise. They do not naturally extend to extremely sparse supervision (e.g., few scribbles) and may struggle with structured artifacts, spatially correlated noise, or uneven background signals commonly observed in real microscopy data. As a result, there remains a critical need for segmentation frameworks explicitly designed for partially annotated, highly noisy, and variable-dimensional biomedical imaging settings.

In this work, we present a novel weakly supervised framework for cell instance segmentation that requires only a small number of user-provided scribbles. Our method leverages a \textit{blind-spot network} training scheme \cite{krull2019noise2void} to define a self-supervised auxiliary objective, motivated by prior studies demonstrating the complementarity between denoising and segmentation tasks \cite{buchholz2020denoiseg,xu2023synergy,wen2024denoising}. Specifically, we introduce an image quantization loss that models foreground and background as arising from a limited number of unimodal intensity distributions, a reasonable prior in many biological imaging scenarios where distinct structures exhibit characteristic intensity levels. The self-supervised component promotes robust foreground–background separation under noisy conditions, while the scribble-based supervision resolves ambiguities between overlapping or touching instances. Unlike existing pipelines that reduce highly multiplexed data to fixed two-channel inputs and require careful tuning of hyperparameters such as object diameter, pixel resolution, or image size, our framework naturally accommodates variable numbers of channels and is agnostic to spatial scale. We demonstrate robust, data-efficient adaptation across diverse imaging representations, achieving accurate instance segmentation at a fraction of the computational and annotation cost required by fully supervised alternatives.

\section{Method}
Our method effectively segments multi-channel noisy images using only a few sparse scribbles denoting nuclei or whole-cell (nucleus + surrounding cytoplasm and membrane compartments) segmentations. We assume that annotators provides annotations indicating foreground regions (e.g., a line inside a cell) and background regions (e.g., rough boundary surrounding the cell). 

Given the variability in image size and scale across acquisitions, we employ a U-net architecture to develop a space-preserving segmentation and classification network. We assume that the background and the target biological instances can be approximated as a mixture of distributions. This assumption introduces a level of quantization to the denoised image, with the resulting reconstruction loss serving to improve the segmentation objective.

Assume that we have a dataset containing $N$ pairs of noisy images and scribbles $\{x^n,y^n\}_{n=1}^N$. Here $x \in \mathbb{R}^{W\times H}$ is a noisy input image and $y \in \{0,1\}^{W\times H\times 2}$ is the corresponding ground truth scribble per-pixel, indicating if a pixel belongs to background (boundaries) or foreground (i.e. cell instance). For notational compactness, we use the sub-index $i$ to refer to the spatial location of a pixel instead of 2-coordinate location $(i,j)$; we also overload the scribble notation $y_i \in \{0,1\}^2$ where $y_i$ sometimes satisfies $ ||y_{i}||_{1} = 1$ (one-hot class encoding where scribbles are present), but otherwise satisfies $||y_{i}||_{1} = 0$ to indicate no scribble is present. Since we consider a low-data regime, we sample a set of $B$ image-scribble pair patches of size $128 \times 128$, denoted as $\{x^b, y^b\}_{b \in B}$, from the available $N$ pairs, creating a total of $N*B$ patches for training. Validation patches are drawn from the same training set, ensuring minimal overlap with the training patches.

\begin{figure*}[t!]
\begin{center}
\includegraphics[width=1.0\linewidth]{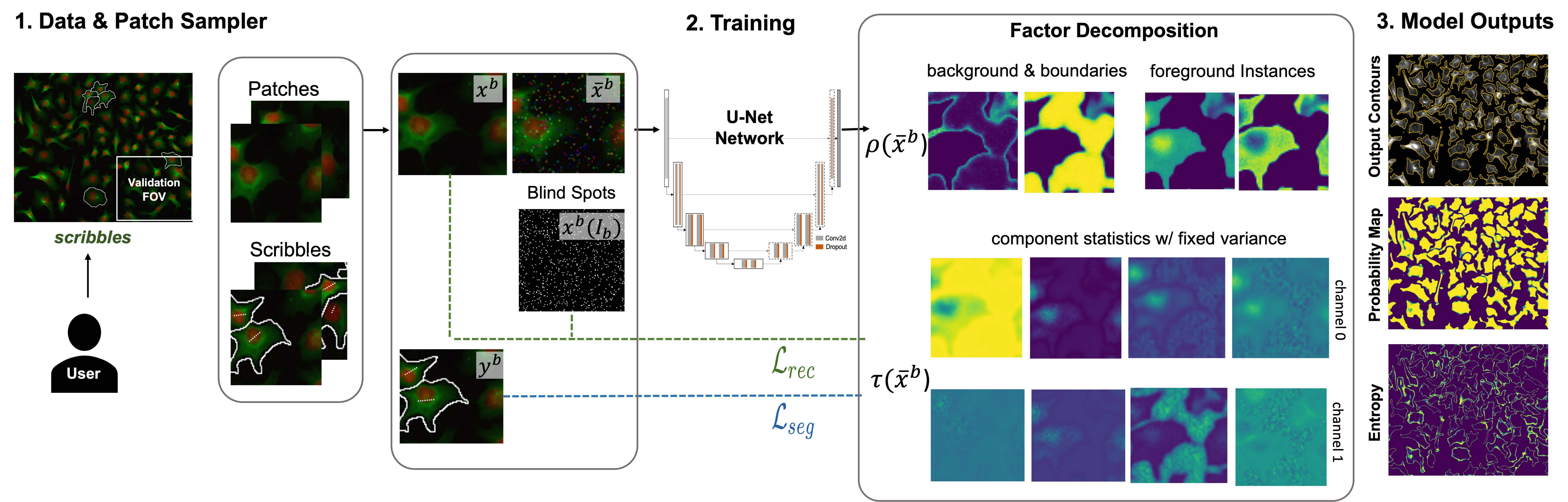}
\end{center}
\caption{Overview of ImPartial training pipeline on multiplex data. This example is modeled with M=4 components ($m_{0}$=2 foreground and  $m_{1}$=2 background)} 
\label{fig:pipeline}
\end{figure*}

\textbf{Blind-Spot Network.} 
Following the \textit{blind-spot network} proposed in Noise2\-Void \cite{krull2019noise2void}, for each image patch $x^b$ in our training set of patches $b \in B$, we generate a partial copy $\bar{x}^b$, where a random set of pixels $I_b$ is substituted by random values in the vicinity ($I_b=\{i : \bar{x}^b_i \gets x^b_{j(i)} , j(i)\neq i\}$) i.e. the blind spot patch, \( \bar{x}^b \), assigns random local values to pixel locations in \( I_b \) while preserving original values elsewhere. The imputation patch \( I_b \) is a mask with nonzero values only at imputed pixel locations, sampled at a probability of 0.2.

\textbf{Reconstruction Loss using Gaussian Mixtures.} We model the intensity of a multi-channel image as a mixture of \( M \) components (\( m = 1 \) to \( M \) in Eq. 1), grouped into \( K \) disjoint subsets corresponding to different classes (e.g., background and foreground nuclei, so \( K = 2 \)). This results in \( M = \sum^{K-1}_{k=0} |m_k| \), where \( m_k \) represents the subset of mixture components modeling the intensity of pixels in class \( k \). The sum of mixture components for class \( k \), \(\sum_{m \in m_k}\rho^m_{\theta,i}(\bar{x}^b)\),  corresponds directly to its probability.  
The mixture reconstruction loss is back propagated exclusively for imputed pixels \( I_b \), preventing the network from learning the identity.

Formally, we define the reconstruction loss of the batch for mixture model with $m \in M$ components as follows:

\begin{align}
\begin{split}
    \mathcal{L}_{mix}(B,\Theta) &= \sum_{b \in B}\sum_{i \in I_{b}}\ell_{mix}(x^b_i;\bar{x}^b,\Theta),  \\
    \ell_{mix}(x^b_i;\bar{x}^b,\Theta) &= \sum_{m} \rho^m_{\theta,i}(\bar{x}^b)\ln p(x^b_i|\tau^{m}_{\theta',i}(\bar{x}^b))), \\
    \rho^m_{\theta,i}(\bar{x}^b)\ge 0 ,&\sum_{m} \rho^m_{\theta,i}(\bar{x}^b) = 1, \forall i,
    %p(x_i|\bar{x}_{RF(i)})=\sum_{m}p(z^{m}_i|\bar{x}_{RF(i)})p(x_i|\bar{x}_{RF(i)},m)\\
    %p(x_i|\bar{x}_{RF(i)},m) = p(x_i|\tau^{m}(\bar{x}_{RF(i)}))\delta()
\end{split}
\label{eq:recloss}
\end{align}

where $\rho^m_{\theta}(.)$ is a parametric (in $\theta$) function that outputs the cluster membership per pixel, $\tau^{m}_{\theta'}(.)$ computes the sufficient static of the distribution of the $m$-th component. The dimensions of $\rho^m_{\theta}(.)$ and $\tau^{m}_{\theta'}(.)$ are therefore $W \times H \times M$, and $W \times H \times (M \times Z)$ respectively, where $Z$ is the number of sufficient statistics for the component distribution (e.g., $Z=1$ for a Gaussian distribution with fixed variance). Note that each pixel is modeled as coming from one of the $m$ component's distribution, with probability $\rho^m_{\theta}(.)$, as opposed to modeling the pixel as a mixture of distributions, like would be the case for a Gaussian mixture model.

The components statistics $\tau^{m}_{\theta'}(.)$ are a function of the observed patch $\bar{x}^b$, and therefore present a dynamic behavior that may be learned from data using a neural network. This allows the adjustment to non-homogeneous images. Both $\rho^m_{\theta}(.)$ and $\tau^{m}_{\theta'}(.)$ are implemented as output channels of our U-Net network, we use the notation $\Theta$ to refer to the entire network parameters, which are partly shared across $\theta$ and $\theta'$. 

\textbf{Scribble Loss.} Given an image patch $x^b$ and its corresponding scribbled image $y^b$, we denote the set of annotated pixels of class $k \in \{0,1\}$ as $S_{b,k} = \{i: y^b_{i,k}=1\}$. Here, $k=0$ represents the background class, while $k=1$ denotes biological instances. We assume each class $k$ is associated with $m_k$ distinct components in the mixture defined in Equation \ref{eq:recloss}, meaning that $M = m_{0} \cup m_{1}$ and $m_{0} \cap m_{1} = 0$. Based on the provided scribbles for background and instances, we propose the following segmentation loss.
\begin{align}
    \begin{split}
    \mathcal{L}^k_{seg}(B,\Theta) &= \sum_{b \in B}  \ell^k_{seg}(y^b,\bar{x}^b;\Theta)/\sum_{b \in B}|S_{b,k}|,
    % \\\ell^k_{seg}(y^b_{i};\bar{x}^b,\Theta) &= \sum_{i \in S_{b,k}} [y^b_{i,k}-\sum_{m \in m_k}\rho^m_{\theta,i}(\bar{x}^b)]^2.
     \\\ell^k_{seg}(y^b_{i};\bar{x}^b,\Theta) &= \sum_{i \in S_{b,k}} -log \Big(\sum_{m \in m_k}\rho^m_{\theta,i}(\bar{x}^b)\Big).
    \end{split}
\label{eq:scribbleloss}
\end{align}

Note that this loss is equivalent to a class-balanced cross entropy loss on the pixels where class labels are available, $y^b_{i,k} =1 $ for $k \in \{0,1\}$, and their associated membership class $\sum_{m \in m_k}\rho^m_{\theta,i}(\bar{x}^b)$ (computed as a sum over class specific component labels); class balancing is achieved by dividing the per-class score over the number of annotated scribbles ($\sum_{b \in B}|S_{b,k}|$). The combination of this objective with the mixture loss enforces regularity on the cluster assignments and therefore on the component statistics $\tau^{m}_{\theta'}(.)$. The dual perspective is that we encourage our recovered class labels to be consistent with the reconstruction objective defined for the mixture loss.

\textbf{Joint loss.} We combine the quantization reconstruction mixture loss and the scribble segmentation loss that makes use of the available scribbles into a single training loss. Given a batch of images of size $B$, we define the following joint objective:
\begin{align}
\begin{split}
    % \min\limits_{\Theta}\sum_{k \in K} \lambda_k %\mathcal{L}^k_{seg} + (1-\sum_{k \in K} %\lambda_k)\mathcal{L}_{mix}.
    \min\limits_{\Theta}\sum_{k \in \{0,1\}} (1-\lambda) \mathcal{L}^k_{seg} + \lambda\mathcal{L}_{mix}.
\end{split}
\label{eq:jointloss}
\end{align}
Note that here $\lambda$ represents the weight given to the regularization loss $\mathcal{L}_{mix}$. %We choose a small budget $(1-\sum_{k \in K}\lambda_k) \in [0.01,0.1]$.}%

\textbf{Extension to multiple channels.} We extend the framework to multiple channels (input image $x \in \mathbb{R}^{W\times H \times C}$) in the natural way by computing the mixture (quantization) statistics jointly across all channels $C$. Across all experiments, each component $M$ in the mixture is modeled as a fixed-variance Gaussian with mean $\tau^m \in \mathbb{R}^C$. The component mask $\rho^m_\theta(\cdot)$ is therefore shared across channels. Since we assume the scribbled mask is related to objects that appear as foreground components on at least one channel, we can still compute the scribble mask by summing over the foreground components of the multi-channel mixture mask. The joint loss therefore remains unchanged.

\section{Results and Discussion}

We compared ImPartial with SOTA algorithms on three widely used, expert-annotated public datasets for multiplexed imaging. The ground truth annotations were randomly sampled to generate the required scribbles.

\noindent
\textbf{TissueNet.} This is a 2-channel dataset \cite{greenwald2022whole} containing a nuclear (e.g. DAPI) and a cytoplasmic or membrane marker (e.g. Pan-Keratin). It is collected from different imaging platforms and contains different tissue types and disease states. We used the originally released TissueNet 1.0 dataset with their train/test splits: train - 2600 (512x512), test - 1250 (256x256).

\begin{table*}[t!]
\caption{Quantitative comparison of SOTA methods on TissueNet data \cite{greenwald2022whole}.}
\label{table:tissuenet}
\setlength\tabcolsep{6 pt}
\small
\centering
    \begin{tabular}{l|c|c|c|c|c|c}
    % \hline

    \hline
        \textbf{Method} & \textbf{Scribbles} & \textbf{mIOU} & \textbf{mDice} & \textbf{AP} & \textbf{mAP$_{0.5:0.95}$} & \textbf{F1$_{0.5}$} \\ 
        % \hline  
        \bottomrule
        Mesmer \cite{greenwald2022whole} & 100\% & 0.7017 & 0.7910 & 0.7385 & 0.4341 & 0.8403 \\ \hline

        Cellpose \cite{pachitariu2022cellpose}  & 100\% & 0.7568 & 0.8449 & 0.7494 & 0.4461  & 0.8333 \\ 
         & 10\% & 0.6952 & 0.7934 & 0.6736 & 0.3502 & 0.8055 \\ \hline

        Cellotype \cite{pang2024cellotype}  & 100\% & 0.7044 & 0.7968 & 0.7416 & 0.4180 & 0.8365 \\ \hline
        CellSAM \cite{Marks2025-xd}  & 100\% & 0.6915 & 0.7858 & 0.7363 & 0.4102 & 0.8391 \\ \hline
        
        ImPartial  & 10\% & 0.7486 & 0.8476 & 0.7290 & 0.3681 & 0.8261 \\ 

          & 20\% & 0.7304 & 0.8426 & 0.7176 & 0.3720 & 0.8208  \\ 

          & 50\% & 0.7490 & 0.8482 & 0.7313 & 0.3856 & 0.8305 \\ 
        
          & 100\% & 0.7475 & 0.8493 & 0.7341 & 0.4031 & 0.8283 \\ \hline
        
    \end{tabular}
\end{table*}

\begin{figure*}[t!]
\begin{center}
\includegraphics[width=1.0\linewidth]{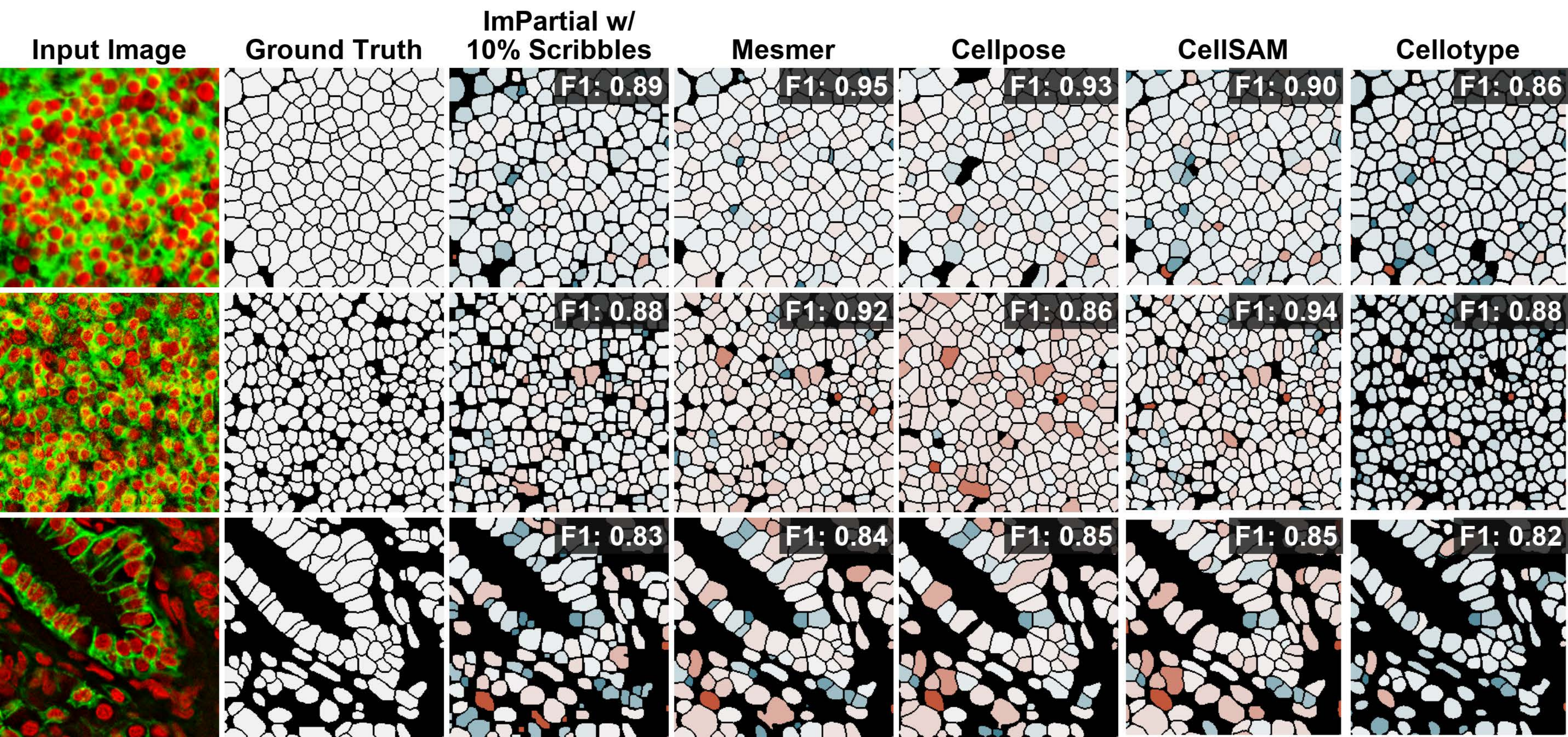}
\caption{Visual comparison on TissueNet dataset results with ground truth, under (red) and over (blue) segmentation predictions for results in Table \ref{table:tissuenet}.}
\end{center}
\end{figure*}

\begin{figure*}[t!]
\centering
\includegraphics[width=0.78\linewidth]{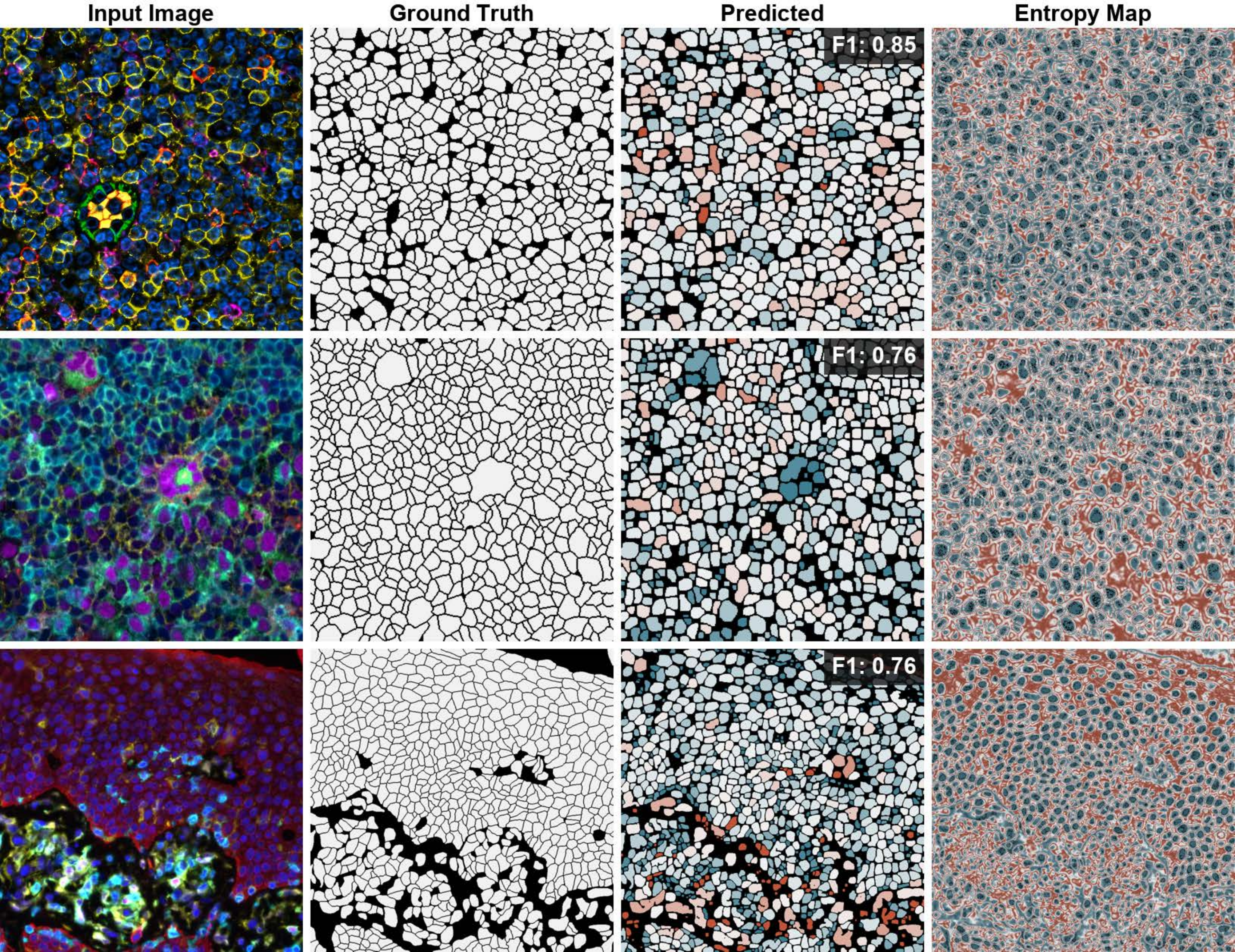}

\vspace{0.5em}
\setlength{\tabcolsep}{7pt}
    \begin{tabular}{l|c|c|c|c|c}
    % \hline
    \hline
        \textbf{Method} & \textbf{Scribbles} & \textbf{mIOU} & \textbf{mDice} & \textbf{AP} & \textbf{F1$_{0.5}$} \\ 
        % \hline  
        \bottomrule
        \hline
        ImPartial & 10\% & 0.5241 & 0.6568 & 0.3841 & 0.6957 \\
        ~ & 20\% & 0.5582 & 0.6892 & 0.4411 & 0.7189 \\ 
        ~ & 50\% & 0.5835 & 0.7105 & 0.5018 & 0.7442 \\ \hline
    \end{tabular}

\hfill
\caption{(Top) Quantitative results on 6-channel CPDMFCI dataset with ground truth masks, under (red) and over (blue) segmentation predictions, and entropy. First row shows lymph node/normal tissue from CODEX platform with 6 channels: CD8 (Red), CD20 (Magenta), CD21 (Cyan), CD31 (Green), CD45RO (Yellow), DAPI (Blue). Second row shows lymph node/Hodgkin's lymphoma tissue from VECTRA platform with 7 channels: CD8 (Red), CD20 (Magenta), CD21 (Cyan), CD31 (Green), CD45RO (Yellow), DAPI (Blue). Third row shows skin/cutaneous T-cell lymphome from Zeiss platform with 5 channels: PanCK (Red), PD-L1 (Green), CD3 (Cyan), Foxp3 (Magenta), DAPI (Blue). (Bottom) Quantitative comparison using different annotation budget.}
\label{fig:cpdmi}
% \end{center}
\end{figure*}

\noindent
\textbf{CPDMFCI.} This dataset \cite{aleynick2023cross} includes annotations on multiplex images from three fluorescent imaging platforms and over 40 antibody markers, offering both whole-cell and nuclear annotations for granular segmentation of cellular compartments. To accommodate platform differences, annotations span tissues acquired through sequential immunofluorescence with unmixing (Akoya Vectra 3.0), sequential IF with narrowband capture (Ultivue InSituPlex with Zeiss Axioscan), and cyclical IF with narrowband capture (Akoya CODEX). We selected 81 images (66 train, 15 test) as 400×400 pixel crops from the larger multiplex image set. See results in Fig. \ref{fig:cpdmi}, note that existing methods do not support 3+ channels.

\begin{figure}[ht!]
\centering
% \begin{minipage}{\textwidth}
% \centering
% \end{minipage}
\includegraphics[width=0.87\linewidth]{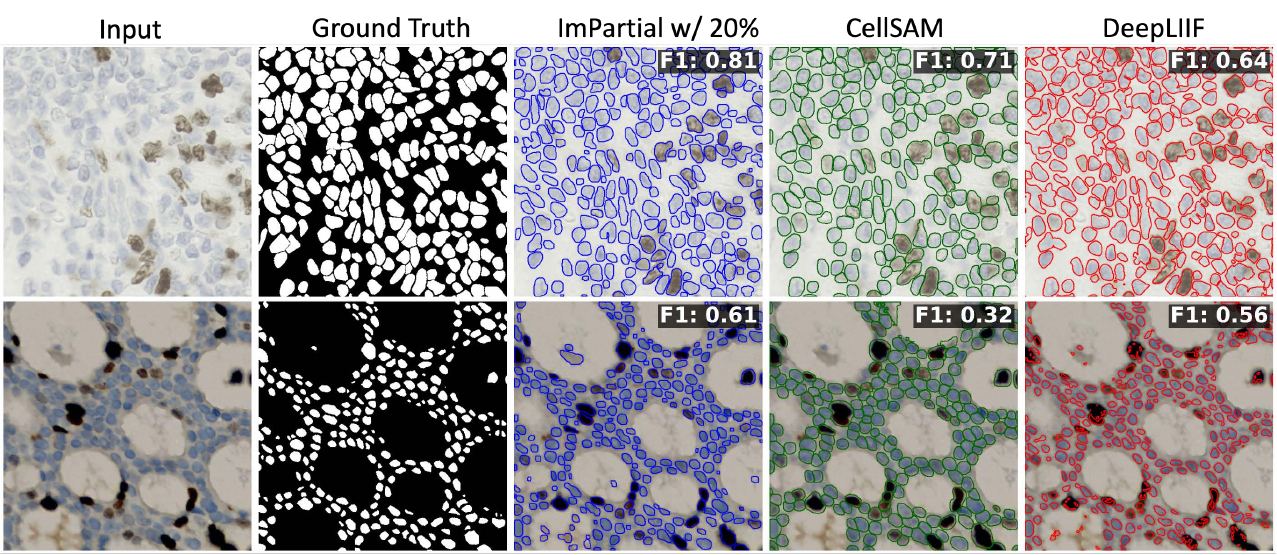} 

\vspace{0.5em}
\setlength{\tabcolsep}{8pt}
    \begin{tabular}{l|c|c|c|c|c}
    \hline
        \textbf{Method} & \textbf{Scribbles} & \textbf{mIOU} & \textbf{mDice} & \textbf{AP} & \textbf{F1$_{0.5}$} \\ 
        % \hline  
        \bottomrule
        \hline 
        CellSAM \cite{Marks2025-xd} & 100\% & 0.4444 & 0.5423 & 0.4397 & 0.5928 \\ \hline
        DeepLIIF \cite{ghahremani2022deep} & 100\% & 0.5841 & 0.6967 & 0.5010 & 0.6570 \\ \hline
        ImPartial & 10\% & 0.6287 & 0.7450 & 0.5394 & 0.6875 \\
        ~ & 20\% & 0.6338 & 0.7488 & 0.5633 & 0.7095 \\ \hline
    \end{tabular}

\caption{Qualitative results (top) on IHC dataset and metrics performance comparison (bottom).}
\label{fig:IHC}
\end{figure}

\noindent
\textbf{IHC.} The publicly available DeepLIIF dataset \cite{ghahremani2022deep}, which consists of co-registered images of IHC, hematoxylin, and multiplex immunofluorescence (mpIF) channels (DAPI, Lap2, and Ki67) acquired from lung and bladder tissue sections. All slides were scanned using a ZEISS Axioscan system. The multi-modal images were preprocessed through scaling and affine co-registration to the corresponding fixed IHC images (512 × 512), 1,975 images were used for training and 391 images testing. Performance comparison is shown in Fig. \ref{fig:IHC}. 

\noindent
\textbf{Evaluation Metrics.} For quantitative evaluation and comparison with other methods, we compute Instance Segmentation metrics such as mean intersection over union (mIOU), Average Precision (AP), F1-score at 0.5 IoU  threshold as implemented in the Stardist \cite{weigert2020} and Cellpose \cite{pachitariu2022cellpose} libraries.

\noindent
\textbf{Ablation Study.} 
We performed an ablation study by first removing the scribble loss followed by removal of the reconstruction loss Table \ref{table:ablation}. When the scribble loss is removed, the method defaults to Noise2Void \cite{krull2019noise2void} and in that case, it is difficult for the model to separate touching or overlapping cells. For the latter case where we removed the reconstruction loss, we show that the method struggles on TissueNet images, especially the ones acquired from the MIBI platform, which tend to have significant noise in the background. These results show that using both loss functions is necessary to obtain optimal results.

\begin{table*}[t]
\caption{Ablation study analysis on TN dataset with 10\% scribbles.} 
\label{table:ablation}
\setlength\tabcolsep{6 pt}
\small
\centering
    \begin{tabular}{l|l|c|c|c|c}

    \hline
        \textbf{Dataset} & \textbf{Loss function} & \textbf{mIOU} & \textbf{mDice} & \textbf{AP} & \textbf{F1$_{0.5}$} \\ 
        % \hline  
        \bottomrule
         TN & Reconstruction Loss Only & 0.0533 & 0.0863 & 0.0051 & 0.0480 \\ 
         & Scribble Loss Only & 0.6952 & 0.7934 & 0.6736 & 0.8055 \\ 
         & Joint Loss & 0.7486 & 0.8476 & 0.7290  & 0.8261 \\ \hline
        
        TN MIBI & Scribble Loss Only & 0.6618 & 0.7665 & 0.6201 & 0.7904 \\ 
         & Joint Loss & 0.7043 & 0.8037 & 0.6796 & 0.8203  \\ \hline
    \end{tabular}
\end{table*}

% \noindent
As an extension to our fully-automated pipeline, we developed an inference stage ensemble approach with Monte Carlo dropout to compute the epistemic uncertainty to find low-performing regions and guide user, to facilitate iterative refinement of model predictions. For example, in Table \ref{table:ablation_iterative} we show that, providing scribbles at higher entropy regions, achieves higher performance with just 4.5\% scribbles than directly giving 5\% scribbles.

\begin{table}[t!]
\setlength\tabcolsep{6 pt}
\small
\centering
    \caption{Iterative ablation on TN data demonstrates that selecting scribbles more intelligently with uncertainty guidance outperforms the baseline of equivalent number of scribbles in the first iteration.}
    \label{table:ablation_iterative}
    \footnotesize
    \begin{tabular}{l|l|c|c|c|c}
    \hline
        ~ & \textbf{Scribbles} & \textbf{mIOU} & \textbf{AP} & \textbf{F1$_{0.5}$} & \textbf{mAP$_{0.5:0.95}$} \\ 
        \bottomrule
        
        iter1 & 0.5\% & 0.6558 & 0.6122 & 0.7777 & 0.3242 \\ \hline
        iter2 & 4.5\% & {0.6896} & {0.6856} & {0.8267} & {0.3739} \\ \hline \hline
        iter1 & 5\% & 0.6761 & 0.6523 & 0.8051 & 0.3414 \\ \hline
        % ~ & 10\% & 80\% & 0.6794 & 0.6724 & 0.8167 & 0.6784 & 0.3680 \\ \hline
        % ~ & 100\% & 80\% & 0.6947 & 0.69750 & 0.8282 & 0.6930 & 0.39207 \\ \hline
    \end{tabular}
\end{table}

\noindent
\textbf{Training details.} We use a U-Net with depth $4$, $64$ initial feature maps, and convolution kernel size of $3$. We use a batch size of size $64$, trained for $200$ epochs and Adam optimizer method with learning rate $4 \times 10^{-4}$, stopping criteria was based on validation performance on the joint training loss. For modeling the cluster components as fixed-variance Gaussian distributions with parametric mean, we used a total of M=4 clusters, 2 each modeling the background and foreground. All the experiments reported in the paper were run for single iteration using 3 different seeds, the standard deviation was $<$ 0.006 across all metrics.

\noindent
\textbf{Acknowledgments.} This project was supported by NIH R37CA295658, MSK Technology Development Fund, and MSK Cancer Center Support Grant/Core Grant (P30 CA008748).

% \bibliography{refs}

% %
% % ---- Bibliography ----
% %
% % BibTeX users should specify bibliography style 'splncs04'.
% % References will then be sorted and formatted in the correct style.
% %
% % \bibliographystyle{splncs04}
% % \bibliography{mybibliography}




% LNCS Homepage, \url{http://www.springer.com/lncs}, last accessed 2023/10/25
% \end{thebibliography}
\end{document}